\documentclass{article}
\usepackage[eatrack]{log_2023}

\usepackage{booktabs}            
\usepackage{multirow}            
\usepackage{amsfonts}            
\usepackage{graphicx}            
\usepackage[english]{babel}
\usepackage{csquotes}
\usepackage{wrapfig}

\newcommand{\framedtext}[1]{%
\par%
\noindent\fbox{%
    \parbox{\dimexpr\linewidth-2\fboxsep-2\fboxrule}{#1}%
}%
}

\usepackage[
      backend=biber,
      style=numeric-comp,
      backref=true,
      natbib=false]{biblatex}
\vspace*{-0.9cm}

\title[Topological Graph Signal Compression]{Topological Graph Signal Compression}

\author[Guillermo Bernárdez et al.]{%
Guillermo Bernárdez$^*$, Lev Telyatnikov$^\dagger$, Eduard Alarcón$^*$, Albert Cabellos-Aparicio$^*$, \\ \normalsize{\textbf{Pere Barlet-Ros$^*$ and Pietro Liò$^\mathsection$}} \\
$^*$Universitat Politècnica de Catalunya, $^\dagger$Sapienza Università di Roma, $^\mathsection$University of Cambridge
}
\vspace{-0.2cm}

\begin{document}

\maketitle

\begin{abstract}
Recently emerged Topological Deep Learning (TDL) methods aim to extend current Graph Neural Networks (GNN) by naturally processing higher-order interactions, going beyond the pairwise relations and local neighborhoods defined by graph representations. In this paper we propose a novel TDL-based method for compressing signals over graphs, consisting in two main steps: first, disjoint sets of higher-order structures are inferred based on the original signal --by clustering $N$ datapoints into $K\ll N$ collections; then, a topological-inspired message passing gets a compressed representation of the signal within those multi-element sets. Our results show that our framework improves both standard GNN and feed-forward architectures in compressing temporal link signals from two real-world Internet Service Provider Networks' datasets --from $30\%$ up to $90\%$ better reconstruction errors across all evaluation scenarios--, suggesting that it better captures and exploits spatial and temporal correlations over the whole graph-based network structure.
\end{abstract}

\vspace{-0.2cm}

\section{Motivation} \label{sec:motivation}

Graph Neural Networks (GNNs)\cite{zhou2020graph} 
have demonstrated remarkable performance in modelling and processing relational data on the graph domain, which naturally encodes binary interactions. Topological Deep Learning (TDL)\cite{papillon2023architectures} methods take this a step further by working on domains that can feature higher-order relations. By leveraging (algebraic) topology concepts to encode multi-element relationships (e.g. simplicial\cite{bodnar2021simplicial}, cell\cite{bodnar2021cellular} and combinatorial complexes\cite{hajij2023topological}), Topological Neural Networks (TNNs) allows for a more expressive representation of the complex relational structure at the core of the data. In fact, despite its recent emergence, TDL is already postulated to become a relevant tool in many research areas and applications\cite{hajij2023topological}, including 
complex physical systems\cite{battiston2021physics}, 
signal processing\cite{barbarossa2020topological,battiloro2023topological}, 
molecular analysis\cite{bodnar2021simplicial,schiff2020characterizing} and social interactions\cite{schaub2020random}.

We argue that the task of data compression can hugely benefit from TDL by enabling to exploit multi-way correlations between elements beyond pre-defined local neighborhoods to get the desired lower-dimensional representations. 
To the best of our knowledge, current Machine Learning (ML) compression approaches mainly rely on Information Theory (IT) and are narrowed to Computer Vision applications\cite{havasi_2021,yang2023introduction}. 
In contrast to that, and inspired by \textit{zfp}\cite{diffenderfer2019error} --the current state-of-the-art lossy compression method for floating-point data, more details in \ref{app:related}--, we propose in this paper a novel TDL framework to \textit{(a)} first detect higher-order correlated structures over a given data, and \textit{(b)} then directly apply TNNs to obtain compressed representations 
within those multi-element sets. 

This work provides evidence supporting that TDL could have great potential in compressing relational data. With the long-term objective of outperforming \textit{zfp}, our current goal is to assess if the proposed framework naturally exploits multi-datapoint interactions --between possibly distant elements-- in a way that makes it more suitable for compression than other ML architectures (even if data comes from the graph domain). 
To do so, we consider the critical problem of traffic storage in today's Internet Service Providers (ISP) networks\cite{almasan2023leveraging}, and set the target to compress the temporal per-link traffic evolution --Figure \ref{fig:use}-- for two real-world datasets extracted from \cite{orlowski2010sndlib} (more details and motivation of this use case provided in \ref{app:use_case}). Once the original link-based signal is divided into processable temporal windows, we benchmark our method against a curated set of GNN-based architectures --and a Multi-Layer Perceptron (MLP)-- properly designed for compression as well. Obtained results clearly suggest that our topological framework defines the best baseline for \textit{lossy neural compression}.

\newpage
\section{Methodology}
\begin{figure}[!t]
\centering
    \includegraphics[width=\columnwidth]{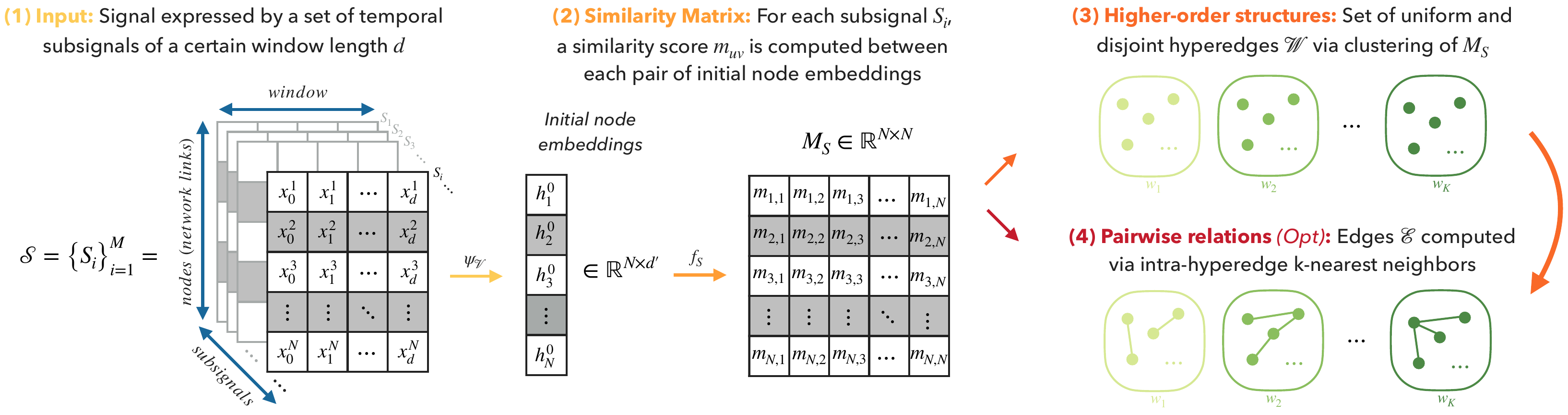}
  \caption{Topology Inference Module. For each subsignal $S_i \in \mathcal{S}$, it outputs a topological object $\mathcal{T} = \left( \mathcal{V}, \mathcal{E}, \mathcal{W} \right)$ determined by $K$ disjoint hyperedges.}
  \label{fig:topology}
\end{figure}

This section describes the proposed Topological (Graph) Signal Compression framework, which is divided into the following three primary modules:

\paragraph{1) Topology Inference Module.} The first stage of the proposed model infers the computational topological structure --both pairwise and higher-order relationships-- from the data measurements. 
In general, the framework assumes to have a set $\mathcal{S}$ of $M$ signals, $\mathcal{S}=\{ S_i \}_{i=1}^{M}$, where $S_i$ consists of $N$ vector-valued measurements $x_j$ of a pre-defined dimension $d$, i.e. $S_i = \{ x^i_j \}_{j=1}^{N}$, $x^i_j \in \mathbb{R}^d$. Thus, the pipeline that we describe as follows (see also Figure \ref{fig:topology}) is independently applied to every signal $S_i$.

\noindent\textbf{Similarity Matrix:} The initial signal $\{ x_j \}_{j=1}^{N}$ \footnote{For the sake of simplicity, and as abuse of notation, we will avoid writing the superscript $i$ when referring to the measurements of a generic signal $S_i \in \mathcal{S}$.} is encoded with a MLP into an embedding space $h_j^0 = \psi_{\theta_{\mathcal{V}}}(x_j) \in \mathbb{R}^{d'}$, $\forall j \in \{1, \dots, N\}$. Next, we compute the pairwise similarity matrix $M_S = (m_{uv}) \in \mathbb{R}^{d' \times d'}$ where $m_{uv} := f_S\left(h^0_u,h^0_v \right)$ and $f_S : \mathbb{R}^{d'} \times \mathbb{R}^{d'} \rightarrow \mathbb{R}$ is a similarity function.

\noindent\textbf{Higher-order Relationships:} 
We use clustering techniques on the similarity matrix $M_{S}$ to deduce $K$ higher-order structures, over which a Topological Message Passing pipeline --see next module \ref{subsec:compression_mp})-- performs the compression. In fact, the idea is to compress the signal within the inferred multi-element sets and encode compressed representations of the data into the final hidden states of these hyperedges. 
Therefore, the number of higher-order structures $K$ is desired to be considerably lower than the number $N$ of datapoints ($K \ll N$); we design the following \textit{clustering} scheme for this purpose:

\begin{enumerate}
    \item The number of hyperedges are defined as $K = \lfloor N/p \rceil $, where $p$ is a hyperparameter that identifies the maximum hyperedge length. 
    
    \item For every row in the similarity matrix $M_S$, we extract the top $p-1$ highest entries and calculate their sum. We then select the row that corresponds to the highest summation value. This chosen row becomes the basis for forming a hyperedge as we gather the indices of the $p-1$ selected columns along with the index of the row itself. Then the gathered indices are removed from the rows and columns of the similarity matrix $M_S$, obtaining a reduced $\hat{M}_S \in \mathbb{R}^{(d'-p)\times(d'-p)}$.
    
    \item Previous step 2 is repeated with subsequent $\hat{M}_S$ until $K$ disjoint hyperedges are obtained.\footnote{When $N/p$ is not an even division, at some point of the process the ranking starts considering the row-wise $p-2$ higher entries to form $p-1$-length hyperedges, so that at the end a total of $K= \lfloor N/p \rceil$ hyperedges of lengths $p$ and $p-1$ are obtained; see \ref{app:our_implementation} for further details.}
\end{enumerate}
On the other hand, the choice of the similarity function becomes a crucial aspect for the compression task. Supported by our early experiments (see Section \ref{sec:results}), our framework makes use of the \textbf{Signal to Noise Ratio (SNR)} distance metric presented in \cite{yuan2019signal}, proposed in the context of deep metric learning as it jointly preserves the semantic similarity and the correlations in learned features\cite{yuan2019signal}.

\noindent\textbf{Pairwise relationships:} Besides higher-order structures, our framework can optionally leverage graph-based relational interactions, either \textit{(i)} by considering the original graph connectivity if it is known, or \textit{(ii)} by inferring the edges via the similarity matrix as well --by connecting each element with a subset of top $k$ row-based entries in $M_S$. 
In our experiments only intra-hyperedge edges have been considered to keep the inferred higher-order structures completely disjoint from each other. 

\paragraph{2) Compression Module via Topological Message Passing}\label{subsec:compression_mp}
We implemented two topological Message Passing (MP) compression pipelines, named \textbf{SetMP} and \textbf{CombMP}. \textbf{SetMP} is a purely set-based architecture that operates only over hyperedges and nodes; more details in \ref{app:method}. In this section we describe  \textbf{CombMP}, our most general architecture that leverages the three different structures (nodes, edges, hyperedges) in a hierarchical way,\footnote{Edges and hyperedges are distinguished because, analogously to recent Combinatorial Complexes (CCC) models\cite{hajij2023topological}, edges can hierarchically communicate with hyperedges if they are contained in them; in fact, the name \textbf{CombMP} relates to these general topological constructions (more details in Appendix \ref{app:related}).} and can be seen as a generalisation of \textbf{SetMP}.

\begin{figure}[!t]
\centering
    \includegraphics[width=\columnwidth]{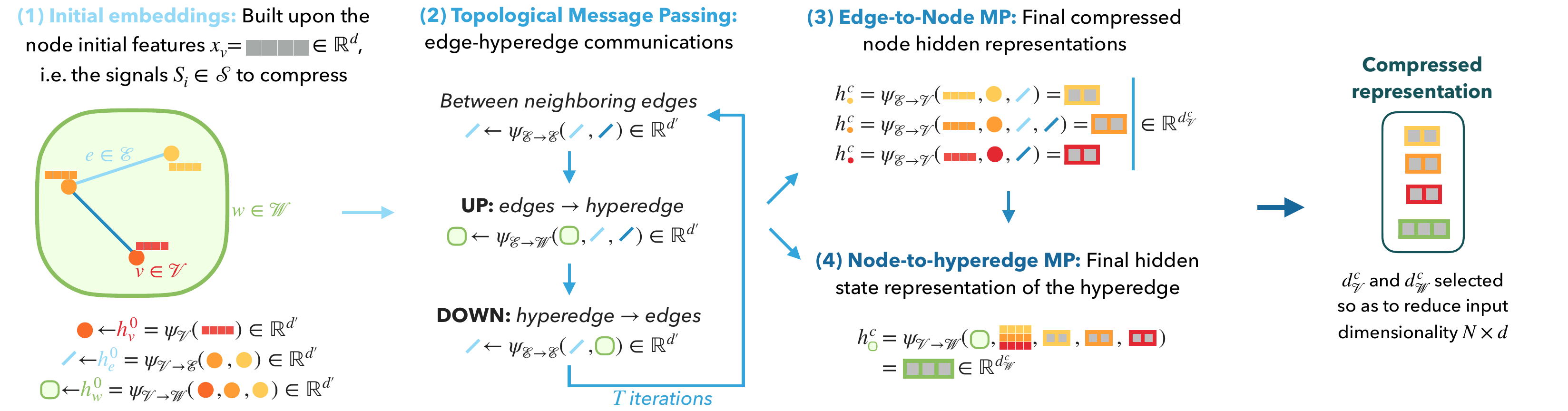}
  \caption{Compression Module workflow for the \textbf{CombMP} architecture; it is independently applied to each hyperedge $w\in\mathcal{W}$ of the inferred topological object $\mathcal{T} = \left( \mathcal{V}, \mathcal{E}, \mathcal{W} \right)$. The arguments of the functions visually represent either initial node features or the corresponding element embedding.}
  \label{fig:compression}
\end{figure} 

For a given signal $S_i$ and its corresponding initial embeddings $\{ h_j^0 \}_{j=1}^{N}$, our model operates over a topological object $\mathcal{T} = \left( \mathcal{V}, \mathcal{E}, \mathcal{W} \right)$ where $\mathcal{V}$ denotes the set of elements or nodes, $|\mathcal{V}|=N$; $\mathcal{E}\in \mathcal{V} \times \mathcal{V}$ represent the set of edges; and $\mathcal{W} \in (\mathcal{V} \times \dots \times \mathcal{V})$ the set of hyperedges.
The compression pipeline (visualized in Figure \ref{fig:compression}) can be described as follows:

\noindent\textbf{Initial embeddings:} First, we generate initial embeddings for the three considered topological structures. For nodes, we use the previously computed embeddings $\{ h_v^0 \}_{v=1}^{N}$. For edges and hyperedges, (learnable) permutation invariant functions are applied over the initial embeddings of the nodes they contain; respectively, $h_e^0 = \phi_{\theta_{\mathcal{E}}} \left( \oplus_{v\in e} h_v^0 \right)$ for each $e\in\mathcal{E}$, and $h_w^0 = \phi_{\theta_{\mathcal{W}}} \left( \oplus_{v\in w} h_v^0 \right)$ for each $w\in\mathcal{W}$. The same dimension $d'$ is used for all initial and intermediate hidden representations.

\noindent\textbf{Edge-Hyperedge Message Passing:} We define a hierarchical propagation of messages between edges and hyperedges. First, neighboring edges communicate to each other to update their representations; denoting the edge neighbors of an edge $e \in \mathcal{E}$ by $\mathcal{N}_{e}^{\mathcal{E}}:= \{ e'=(u,v)\in \mathcal{E} | e' \neq e, u \in e \lor v \in e \}$, its new hidden state becomes 
$h_e^1 = \phi_{\theta_{\mathcal{E}\rightarrow\mathcal{E}}} \left( \oplus_{e'\in \mathcal{N}_{e}^{\mathcal{E}}} \psi_{\theta_{\mathcal{E}\rightarrow\mathcal{E}}} \left( h_e^0, h^0_{e'} \right) \right) $. 
Next, hyperedges also update their hidden states based on the updated edge representations according to 
$h_w^1 = \phi_{\theta_{\mathcal{E}\rightarrow\mathcal{W}}} \left( \oplus_{e\in \mathcal{E}, e \subset w} \psi_{\theta_{\mathcal{E}\rightarrow\mathcal{W}}} \left( h_w^0, h^1_{e} \right) \right) $, for each $w \in \mathcal{W}$. Then the idea is to propagate downwards towards the edges, i.e. from hyperedges to edges, 
$h_e^2 = \phi_{\theta_{\mathcal{W}\rightarrow\mathcal{E}}} \left( \oplus_{w \in \mathcal{W}, e \subset w} \psi_{\theta_{\mathcal{W}\rightarrow\mathcal{E}}} \left( h_e^1, h^1_{w} \right) \right) $; 
and only between edges again. This whole communication process can be iterated $T$ times.

\noindent\textbf{Edge-to-Node Compression:} At this point, we perform a first compression step over the nodes by leveraging the updated edge hidden representations, the initial node embeddings, as well as the original node data as a residual connection. 
Formally, for each node $v\in\mathcal{V}$ we get a compressed hidden representation
$ h_v^c = \phi_{\theta_{\mathcal{E}\rightarrow\mathcal{V}}} \left( \oplus_{e \in \mathcal{E}, v \in e} \psi_{\theta_{\mathcal{E}\rightarrow\mathcal{V}}} \left( x_v, h_v^0, h^t_{e} \right) \right) \in \mathbb{R}^{d_{\mathcal{V}}^c}$. 

\noindent\textbf{Node-to-Hyperedge Compression:} Finally, a second and last compression step is performed over the hypergraph representations, in this case leveraging a residual connection to the original measurements, the previously computed compressed representations of nodes, as well as the updated hidden representations of hyperedges. More in detail, each hyperedge $w\in \mathcal{W}$ obtains its final compressed hidden representation as
$ h_w^c = \phi_{\theta_{\mathcal{V}\rightarrow\mathcal{W}}} \left( \oplus_{v \in \mathcal{V}, v \in w} \psi_{\theta_{\mathcal{V}\rightarrow\mathcal{W}}} \left( x_v, h_v^c, h^t_{w} \right) \right) \in \mathbb{R}^{d_{\mathcal{W}}^c}$. 

\framedtext{The final node and hyperedge states, $\left\{ \{ h_v^c \}_{v\in \mathcal{V}}, \{ h_w^c \}_{w\in \mathcal{W}} \right\}$, encode the compressed representation of a signal $S_i = \{ x_j \}_{j=1}^{N}$. Consequently, the compression factor $r_c$ can be expressed as:
\begin{equation} \label{eq:ratio}
    r_{c} = \frac{N\cdot d_{\mathcal{V}}^c + K \cdot d_{\mathcal{W}}^c}{N \cdot d}
\end{equation}}

\newpage

\begin{wrapfigure}{r}{0.45\textwidth}
    \centering
    \includegraphics[width=0.44\textwidth]{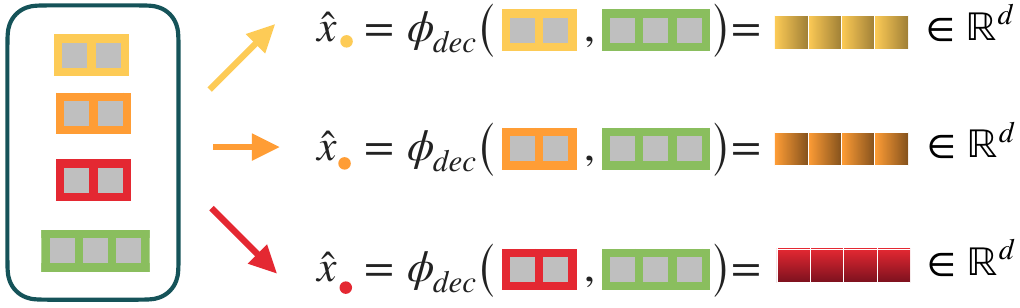}
  \caption{Decompression Module. It is applied over each hyperedge-dependent compressed representation set generated by the Compression Module.
  }
  \label{fig:decompression}
\end{wrapfigure}
\paragraph{3) Decompression Module.}\label{subsec:decompression} This last module learns to reconstruct the original signal of every node through its compressed representation and the final hidden state of the hyperedge it belongs to. 
More formally, for each $v\in\mathcal{V}$ and its corresponding hyperedge $v \in w\in\mathcal{W}$, the reconstructed signal $\hat{x}_v$ is obtained as 
$\hat{x}_v = \phi_{dec} \left( h_v^c, h_w^c \right)$, 
where $\phi_{dec}$ is implemented as a MLP in our framework. 
The whole model is trained end-to-end to minimize the (mean squared) reconstruction error.

\section{Evaluation \& Discussion} \label{sec:results}

\begin{table}[t!]
	\centering
    \caption{Reconstruction Mean Squared Error (MSE) and Mean Absolute Error (MAE) over the test set of the considered datasets for two different compression factors. \textbf{Top:} Methods that leverage the original graph-based network structure. \textbf{Middle:} Methods with no inductive biases. \textbf{Bottom:} Methods that leverage higher-order structures (ours).}
    \label{tab:results}
    \resizebox{1.0\textwidth}{!}{%
	\begin{tabular}{ccccccccc}
		\toprule
		\multirow{3.5}{*}{} & \multicolumn{4}{c}{Abilene} & \multicolumn{4}{c}{Geant}  \\
		\cmidrule(lr){2-5} \cmidrule(lr){6-9}
        & \multicolumn{2}{c}{$r_c = 1 / 3$} & \multicolumn{2}{c}{$r_c = 2 / 3$} & \multicolumn{2}{c}{$r_c = 1 / 3$} & \multicolumn{2}{c}{$r_c = 2 / 3$} \\
        \cmidrule(lr){2-3} \cmidrule(lr){4-5} \cmidrule(lr){6-7} \cmidrule(lr){8-9} 
          & MSE         & MAE        & MSE         & MAE        & MSE         & MAE        & MSE         & MAE        \\
		\midrule
GNN       &$1.95\cdot10^{-2}$&$1.08\cdot10^{-1}$&$1.95\cdot10^{-2}$&$1.08\cdot10^{-1}$&$2.33\cdot10^{-2}$&$1.21\cdot10^{-1}$&$2.32\cdot10^{-2}$&$1.20\cdot10^{-1}$\\
MPNN      &$7.88\cdot10^{-4}$&$1.24\cdot10^{-2}$&$7.92\cdot10^{-4}$&$1.24\cdot10^{-2}$&$8.45\cdot10^{-3}$&$4.13\cdot10^{-2}$&$1.82\cdot10^{-3}$&$2.39\cdot10^{-2}$\\
\midrule
MLP       &$1.04\cdot10^{-3}$&$1.88\cdot10^{-2}$&$9.71\cdot10^{-4}$&$1.80\cdot10^{-2}$&$3.76\cdot10^{-3}$&$3.96\cdot10^{-2}$&$3.62\cdot10^{-3}$&$3.89\cdot10^{-2}$\\
\midrule
SetMP     &\boldmath{$3.22\cdot10^{-4}$}&\boldmath{$8.75\cdot10^{-3}$}&\boldmath{$2.03\cdot10^{-4}$}&\boldmath{$6.80\cdot10^{-3}$}&\boldmath{$6.93\cdot10^{-4}$}&\boldmath{$1.52\cdot10^{-2}$}&\boldmath{$2.90\cdot10^{-4}$}&\boldmath{$1.05\cdot10^{-2}$}\\
CombMP    &$5.81\cdot10^{-4}$&$1.12\cdot10^{-2}$&$3.76\cdot10^{-4}$&$1.06\cdot10^{-2}$&$1.07\cdot10^{-3}$&$1.88\cdot10^{-2}$&$7.04\cdot10^{-4}$&$1.61\cdot10^{-2}$\\
		\bottomrule
	\end{tabular}
    }
\end{table}


\noindent\textbf{Experimental Setup:} For the evaluation, we use two public real-world datasets --Abilene, Geant-- from \cite{orlowski2010sndlib}. They are pre-processed to generate subsignals $S_i$ of network link-level traffic measurements in temporal windows of length $d=10$, to which then a random $60/20/20$ split is performed for training, validation and test, respectively. In this context, our method is compared against:
\begin{itemize}
    \item w we implemented several standard GNN architectures (GCN\cite{kipf2017semi}, GAT\cite{velivckovic2017graph}, GATv2\cite{brody2021attentive}, GraphSAGE\cite{hamilton2017inductive}) to perform signal compression over the network graph topology; we take the best result among them in each evaluation scenario.
    \item \textbf{MPNN}: a custom MP-based GNN scheme --over the original network graph structure as well-- whose modules and pipeline are similar to our proposed topological MPs.
    \item \textbf{MLP}: a feed-forward auto-encoder architecture with no inductive biases over subsignals $S_i$.
\end{itemize}
\textbf{GNN} and \textbf{MPNN} baselines implement a decompression module similar to that of our TDL-based methods. More details about the evaluation and all model implementations are provided in \ref{app:setup}.

\noindent\textbf{Experimental Results:} Table \ref{tab:results} shows the reconstruction error (MSE and MAE) obtained by our framework and the baselines in both datasets for two compression factors, $1/3$ and $2/3$. \textbf{SetMP}, our topological edge-less architecture, clearly performs the best in all scenarios --improving on average by $75\%$ and $48\%$, respectively, the best MSE and MAE obtained by baselines--, followed by our most general \textbf{CombMP} method. As for the baselines, in overall \textbf{MPNN} slightly outperforms \textbf{MLP}, and \textbf{GNN} performs the worst. In addition, a comparison against state-of-the-art \textit{zfp} can be found in \ref{app:zfpvs}. 

\noindent\textbf{Discussion:} These results support our hypothesis that taking into account higher-order interactions 
could help in designing more expressive ML-based models for (graph) signal compression tasks, specially due to the fact that these higher-order structures can go beyond the (graph) local neighborhood and connect possibly distant datapoints whose signals may be strongly correlated (e.g. generator and sink nodes in ISP Networks). 
In that regard, TDL can provide us with novel methodologies that naturally encompass and exploit those multi-element relations. Moreover, it is interesting to see how our set-based architecture outperforms the combinatorial-based one in every scenario, suggesting that intermediate binary connections might add noise in the process of distilling compressed representations. 
Further discussion on future work, focusing on the current limitations of our method and how to possibly address them, can be found in \ref{app:future}. 

\newpage



\section*{Acknowledgements}
This publication is part of the Spanish I+D+i project TRAINER-A (ref.~PID2020-118011GB-C21), funded by MCIN/AEI/10.13039/ 501100011033. This work is also partially funded by the Catalan Institution for Research and Advanced Studies (ICREA), the Secretariat for Universities and Research of the Ministry of Business and Knowledge of the Government of Catalonia, and the European Social Fund.

\printbibliography

\newpage
\appendix
\section{Appendix}

\subsection{Network Traffic Compression} \label{app:use_case}


Computer Network's traffic has significantly increased in recent years\cite{tune2013internet}, specially driven by the development of new applications --such as vehicular networks, Internet of Things, virtual reality, video streaming-- and the advancement of network technology --e.g. the fast improvements in link speed. 
In fact, current Internet Service Providers (ISP) Networks can easily produce hundreds of terabytes of traffic traces per day\cite{roy2015inside}, and this keeps growing.

However, network operators continuously need to store and analyze network traffic data for various network management purposes, including network planning, traffic engineering, traffic classification, anomaly detection or network forensics. With those huge amounts of generated data, the efficient storage of all this information is then becoming a crucial aspect for them\cite{xu2016understanding}.

\begin{wrapfigure}{l}{0.6\textwidth}
    \centering
    \includegraphics[width=0.59\textwidth]{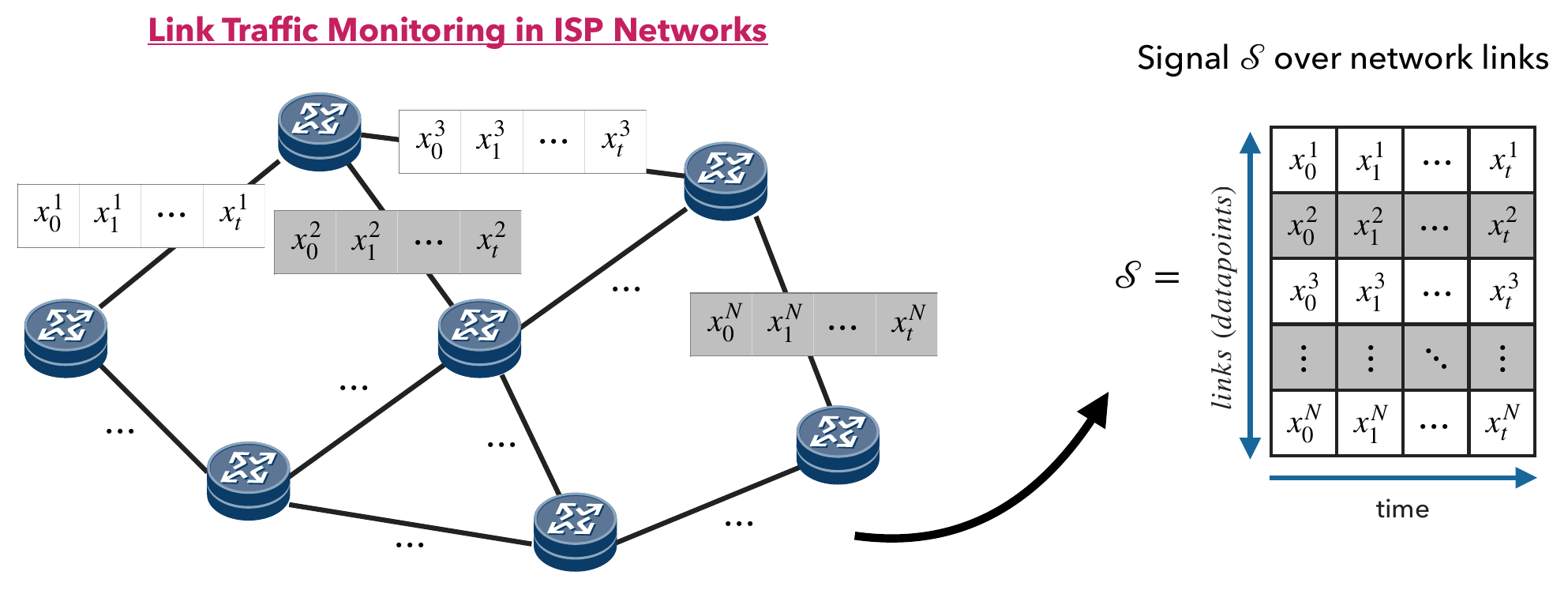}
  \caption{The goal is to compress a signal $\mathcal{S}$ over graph-based ISP Networks representing the temporal evolution of each link utilization.}
  \label{fig:use}
\end{wrapfigure}
This is what motivated us to choose ISP Traffic Compression for testing our proposed method. Not only it provides with complex data naturally represented in the graph domain, but also represents a relevant use case for the networking community. In addition to that, for network management tasks there is a reasonable loss tolerance of the compression, so it would make sense to consider lossy methods such as \textit{zfp}\cite{diffenderfer2019error} or ours. Finally, it is also important to note that there exists public real-world datasets of ISP backbone networks\cite{orlowski2010sndlib} that, despite of their limited network sizes, already reflect complex traffic patterns that may go beyond the provided graph structure (e.g. with distant elements possibly having strong correlations, such as links that are adjacent to generator and sink nodes). Thus, we argue that this use case perfectly serves for our first testing purposes. In particular, in our experiments we consider Abilene and Geant datasets from \cite{orlowski2010sndlib}, which are the ones with the higher number of traffic traces; more details about them in Appendix \ref{app:datasets}.

\subsection{Related Work} \label{app:related}

As already mentioned in Section \ref{sec:motivation}, there already exist ML-based models in the literature that target compression tasks\cite{havasi_2021,yang2023introduction}, but they are mainly entangled to Information Theory concepts used in classical compression algorithms, and applied to Computer Vision domains. Our approach differs from them in these two basic aspects, and has been inspired by \textit{zfp}\cite{diffenderfer2019error}, the state-of-the-art lossy method for floating-point data compression. As detailed in \cite{diffenderfer2019error}, the first step of \textit{zfp} consists in dividing floating matrices or tensors in disjoint blocks of a fixed dimension, which then are independently processed to extract compressed representations. This has obvious similarities with our search of higher-order structures, with the difference that in \textit{zfp} the divisions are totally determined by the input elements' order.

Precisely our proposed topology inference module has some resemblances to that of \cite{xu2022groupnet}, which is also based on grouping entries of an affinity (i.e. similarity) matrix computed over a set of element's neural embeddings. Nevertheless, due to the constraints imposed by the compression task there exists relevant differences between that inference procedure and ours: whereas we design a clustering methodology to get disjoint and uniform-length sets, in \cite{xu2022groupnet} they implement a multi-scale hyperedge forming pipeline where each node ends up belonging to an arbitrary number of higher-order structures --and not only among different hyperedge degrees, but also within the same scale.

Regarding our topological-inspired MP architectures, we highlight the paper of \cite{hajij2023topological} on Combinatorial Complexes (CCC) and the survey \cite{papillon2023architectures} on TDL architectures. Our most general method --CombMP-- was conceptualized before the publication of these works, but we note that it can be formalized in terms of CCCs' notation by considering nodes as 0-cells, edges as 1-cells, and hyperedges as 2-cells; it is due to this fact that we called it CombMP, which stands for \textit{Combinatorial Message Passing}. SetMP, on the other hand, belongs to the hypergraph family of TDL architectures according to the classification of \cite{papillon2023architectures}. More precisely, it can be linked to DeepSets architectures\cite{zaheer2017deep}, which is why it received that name.

Finally, a special mention should be given to \cite{almasan2023leveraging}, which also leverages a ML model to specifically perform traffic compression on ISP networks. However, authors of this work implement a spatio-temporal GNN that acts as a \textit{predictor} of a traditional lossless compression method (in this case, Arithmetic Coding), which defines a totally different conceptual approach.

\subsection{Compression via Topological Message Passing (SetMP)} \label{app:method}

\begin{figure}[!t]
\centering
    \includegraphics[width=\columnwidth]{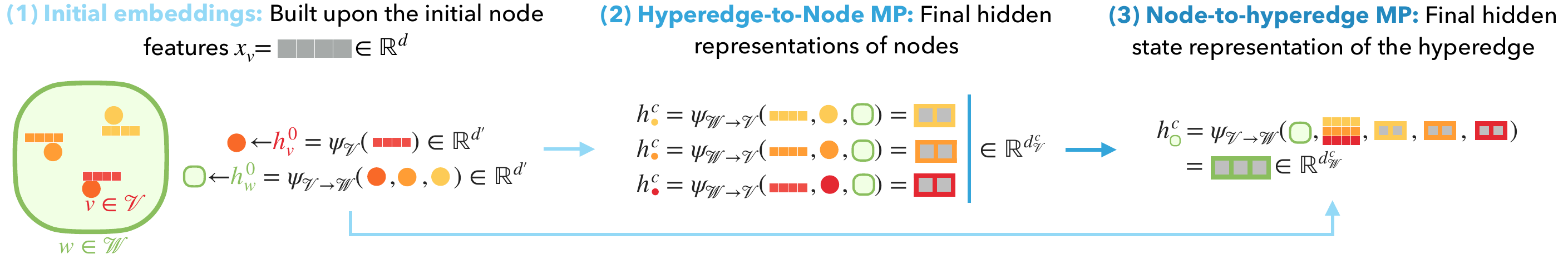}
  \caption{Compression Module workflow for the SetMP architecture; it is independently applied to each inferred hyperedge $w\in\mathcal{W}$. The arguments of the functions visually represent either initial node features or the corresponding element embedding.}
  \label{fig:setmp}
\end{figure} 

In this section we describe the topological MP pipeline of \textbf{SetMP} (Figure \ref{fig:setmp}). In contrast to \textbf{CombMP}, this architecture disregards binary connections and consequently operates over a topological object $\mathcal{T} = \left( \mathcal{V}, \mathcal{W} \right)$, where again $\mathcal{V}$ denotes the set of $N$ nodes and $\mathcal{W} \in (\mathcal{V} \times \dots \times \mathcal{V})$ the set of $K$ inferred hyperedges.
We describe the differences in the compression pipeline in this scenario:

\paragraph{Initial embeddings:} Initial embeddings for nodes and hyperedges are generated in the same way: $\{ h_v^0 \}_{v=1}^{N}$ for the nodes, and $h_w^0 = \phi_{\theta_{\mathcal{W}}} \left( \oplus_{v\in w} h_v^0 \right)$ for each $w\in\mathcal{W}$ for the hyperedges (both of them with dimension $d'$).

\paragraph{Hyperedge-to-Node Compression:} Without edges as intermediaries, we directly perform the node compression, in this case based on both the initial node and hyperedge embeddings and the original signal $S_i$. Formally, for each node $v\in\mathcal{V}$ we get a compressed hidden representation
$ h_v^c = \phi_{\theta_{\mathcal{W}\rightarrow\mathcal{V}}} \left( \oplus_{w \in \mathcal{W}, v \in e} \psi_{\theta_{\mathcal{W}\rightarrow\mathcal{V}}} \left( x_v, h_v^0, h^0_{w} \right) \right) \in \mathbb{R}^{d_{\mathcal{V}}^c}$. 

\paragraph{Node-to-Hyperedge Compression:} The second and last compression step over the hyperedge representations is exactly the same as in the \textbf{CombMP} --except for the fact that now the considered hyperedge hidden states have not been updated. Therefore, each hyperedge $w\in \mathcal{W}$ obtains its final compressed hidden representation as
$ h_w^c = \phi_{\theta_{\mathcal{V}\rightarrow\mathcal{W}}} \left( \oplus_{v \in \mathcal{V}, v \in w} \psi_{\theta_{\mathcal{V}\rightarrow\mathcal{W}}} \left( x_v, h_v^c, h^0_{w} \right) \right) \in \mathbb{R}^{d_{\mathcal{W}}^c}$. 

Thus, again the final node and hyperedge representations --$\left\{ \{ h_v^c \}_{v\in \mathcal{V}}, \{ h_w^c \}_{w\in \mathcal{W}} \right\}$-- encode the compressed representation of a signal $S_i = \{ x_j \}_{j=1}^{N}$, and the same expression of the compression factor $r_c$ applies to \textbf{SetMP}. Finally, we note that the decompression module does not vary either.

\subsection{Further Details of Evaluation} \label{app:setup}

In this section we provide more technical details about the datasets and model implementations. Lastly, we also show the performance of \textit{zfp} in the considered evaluation scenarios.

\subsubsection{Datasets} \label{app:datasets}

As stated in Section \ref{sec:results}, the traffic traces of both datasets are publicly available at \cite{orlowski2010sndlib}. After distributing them into links, Abilene dataset contains link-level traffic utilization measurements over 6 months --in intervals of 5 minutes-- for a topology with 12 nodes and 30 directional links. Considering a temporal window of length $d=10$, this results in $N=4,809$ subsignal samples after data cleaning. 
On the other hand, Geant dataset contains analogous measurements for a period of 4 months and a time interval of 15 minutes, in this case for a topology with 22 nodes and 72 directional links. After data cleaning, it has a total of $1,075$ final samples with the same window size. 

The topology structure of both networks is also provided, and we use it in our GNN-based baselines. The aforementioned $60/20/20$ split is performed over these resulting link-based subsignals.

\subsubsection{Implementation of our Proposed Models} \label{app:our_implementation}

Regarding the Topology Inference module, we recall the relevance of the hyper-parameter $p$ that defines the maximum allowed hyperedge length, and which also defines the number of those hyperedges by $K = \lfloor N/p \rceil $, being $N$ the total number of datapoints. In particular, in our implementation we try to form $K$ $p$-uniform disjoint hyperedges if possible, but otherwise build a combination of $p-1$ and $p$-uniform disjoint hyperedges --so that every datapoint is contained in one of them. Moreover, we always consider $p>4$ so that $K \ll N$ holds.

As shown in Equation \ref{eq:ratio}, this parameter $p$ together with the dimensions of the final node and hyperedge compressed representations, $d_{\mathcal{V}}^c$ and $d_{\mathcal{W}}^c$, define the compression factor of our method. After some hyperparameter tuning, in our experiments we used $p=8$, $d_{\mathcal{V}}^c=2$ and $d_{\mathcal{W}}^c=10$ for achieving $r_c = 1/3$, and $p=6$, $d_{\mathcal{V}}^c=5$ and $d_{\mathcal{W}}^c=10$ for $r_c = 2/3$; respectively, this resulted in $4$ and $5$ inferred hyperedges for the Abilene dataset, and $9$ and $12$ for Geant.

These parameters apply to both SetMP and CombMP architectures, and in both scenarios we also consider node, edge and hyperedge hidden representations of dimension $d'=20$, double the the dimension $d$ of node signals. All message and update functions, $\psi_{\theta_\cdot}$ and $\phi_{\theta_\cdot}$, are implemented as MLPs, and permutation-invariant aggregator functions $\oplus$ consist of the concatenation three element-wise operations: mean, max and min. In the case of CombMP, only 1 iteration of topological message passing (Figure \ref{fig:compression}.2) is performed. Finally, we note that the Decompression Module has the same MLP structure in both compression pipelines.

We will publicly release the code in the future extended version of this work.

\subsubsection{Baseline Implementations}

Analogously to what we do with our proposed architectures, we have fine-tuned the involved hyperparameters of all implemented baselines to perform the compression task. Moreover, they follow a similar compression scheme than our proposed TDL methods. We provide more details below:

\paragraph{Graph-based} These methods leverage the original graph-like network structure present in Abilene and Geant datasets. In this case, since the signal is over the edges, we compute the (dual) line graph of the network, so that edges become nodes and are connected between them if they share origin/destination. The idea is then to perform several iterations of message passing over this line graph to get a compressed representation of the original signal in the hidden state of these link-based nodes. The difference between our two graph-based baselines precisely relies on the nature of this message passing:
\begin{itemize}
    \item \textbf{GNN:} Under this name we gather the results of implementing several standard GNN architectures (GCN\cite{kipf2017semi}, GAT\cite{velivckovic2017graph}, GATv2\cite{brody2021attentive}, GraphSAGE\cite{hamilton2017inductive}). In all of these cases, two consecutive message interchanges (with relatively high dimensional hidden states, $64$ and $32$ in our experiments) are performed before a third one gets the desired compressed representation. We have used the available implementations of PyTorch Geometric \cite{fey2019fast} for the convolutional GNN layers, and performed an exhaustive hyperparameter-tuning for each of them (testing different hidden dimensions, aggregations, normalizations, dropout values, number of heads, etc.). As stated in Section \ref{sec:results}, in each dataset/compression ratio scenario we select the best performing model among this set of GNN architectures to perform the evaluation (shown in Table \ref{tab:results}). However, we note that there is not a significant difference in performance among the different GNN models, and within a single model different hyperparameter settings do not result in huge performance variations either; we will further expand this analysis in future work.
    \item \textbf{MPNN:} In this case we implement a custom Message Passing GNN whose pipeline resembles that of the CombMP architecture: edges to edges, edges to nodes, nodes to edges, and a final edge to node communication with a residual connection to the original node signal that performs the compression. In this case the intermediate node and edge hidden states'  dimension is set to $d'=20$, just as in our topological-inspired methods. Notably, the implementation of this baseline follows directly from the topological architectures, restricting everything to the graph domain. As a result, it underwent hyperparameter tuning in exactly the same way as CombMP and SetMP.
\end{itemize}
In both cases, the final hidden states obtained from the last MP step represent the node signal compressed representations. Since the original window-based signals have length $d=10$, we set this final dimension to $4$ and $7$ when benchmarking our method against them for getting compression factors $r_c$ of $1/3$ and $2/3$, respectively. Finally, we note that both graph-based baselines implement a MLP for the decompression task totally analogous to the one of our TDL-based architectures (in this case simply having as input the final node compressed representation).

\paragraph{MLP} We also implemented a MLP auto-encoder architecture that considers all possible connections between all elements of each subsignal $S_i=\{ x_j \}_{j=1}^{N}$. In particular, each of the subsignals is flattened --i.e. $d_{input} = N \cdot d$-- and passed to a feed forward encoder network (with $1024$ and $512$ hidden dimensions and ReLu activation function after our architecture search) that outputs a final compressed representation of the full subsignal with dimension $d_{output} = \lceil r_c \cdot d_{input} \rceil$, being $r_c$ the considered compression factor. A symmetric decoder network reconstructs the signal from that representation.

\subsubsection{Training and Validation Pipeline}

All models are trained for a maximum of $200$ epochs in Abilene dataset and $500$ in Geant using Adam optimizer (with learning rate $0.003$ and epsilon $0.001$) and using batches of $25$ samples if required. The model state corresponding to the best performing iteration over the validation set is selected for the test evaluation, whose results are represented in Table \ref{tab:results} of the main body of the paper.

\subsubsection{Comparison against \textit{zfp}} \label{app:zfpvs}

\begin{table}[t!]
	\centering
    \caption{Reconstruction Mean Squared Error (MSE) and Mean Absolute Error (MAE) over the test set obtained by our best performing architecture (SetMP) and the state-of-the-art method \textit{zfp}.}
    \label{tab:zfp}
    \resizebox{1.0\textwidth}{!}{%
	\begin{tabular}{ccccccccc}
		\toprule
		\multirow{3.5}{*}{} & \multicolumn{4}{c}{Abilene} & \multicolumn{4}{c}{Geant}  \\
		\cmidrule(lr){2-5} \cmidrule(lr){6-9}
        & \multicolumn{2}{c}{$r_c = 1 / 3$} & \multicolumn{2}{c}{$r_c = 2 / 3$} & \multicolumn{2}{c}{$r_c = 1 / 3$} & \multicolumn{2}{c}{$r_c = 2 / 3$} \\
        \cmidrule(lr){2-3} \cmidrule(lr){4-5} \cmidrule(lr){6-7} \cmidrule(lr){8-9} 
          & MSE         & MAE        & MSE         & MAE        & MSE         & MAE        & MSE         & MAE        \\
		\midrule
SetMP     &$3.22\cdot10^{-4}$&$8.75\cdot10^{-3}$&$2.03\cdot10^{-4}$&$6.80\cdot10^{-3}$&$6.93\cdot10^{-4}$&$1.52\cdot10^{-2}$&$2.90\cdot10^{-4}$&$1.05\cdot10^{-2}$\\
\textit{zfp}     &$9.19\cdot10^{-5}$&$7.34\cdot10^{-3}$&$4.02\cdot10^{-7}$&$4.84\cdot10^{-4}$&$1.04\cdot10^{-4}$&$7.83\cdot10^{-3}$&$4.18\cdot10^{-7}$&$4.95\cdot10^{-4}$\\
		\bottomrule
	\end{tabular}
    }
\end{table}

Finally, we extend the evaluation by showing the relative performance of our best behaved TDL-based methodology --SeMP-- with respect to the state-of-the-art lossy compression method \textit{zfp} for the desired precision; see Table \ref{tab:zfp}. As it can be seen, \textit{zfp} gets better reconstruction errors than our model in all scenarios, although we note that differences are considerably lower for smaller (i.e. more challenging) compression factors. Overall, we reckon that these are promising results; our topological models are postulated as strong ML-based baselines for lossy compression, and by addressing some of their current limitations (see Section \ref{app:future}) there may be room for for shortening the gap w.r.t \textit{zfp}.

In particular, we hypothesize that the main limitation of our current method revolves around its topology inference procedure, as it can potentially gather elements that do not necessarily share any correlation --especially at the end of the clustering process, where groups are built in a greedy manner regardless of the actual similarities between their elements. Whereas \textit{zfp} implements a sophisticated method to deal with such uncorrelated elements, our current methodology mainly relies on existing correlations to perform compression; this can lead to a systematic poor reconstruction error of the involved signals. We will delve deeper into this in the extension of this work.

\subsection{Future Work} \label{app:future}

Despite the promising preliminary results obtained, and because of them as well, there are many aspects and limitations of our proposed methodology that are being investigated and will be addressed in future work. The following list summarizes some of the main lines of research:

\paragraph{Topology Inference} Apart from exploring other metrics beyond SNR, it would be interesting to consider more flexible clustering approaches that could dynamically adapt the length/number of the inferred higher-order structures (e.g. by defining a Reinforcement Learning pipeline to perform the division, or adapting a solution like the Differentiable Cell Complex Module\cite{battiloro2023latent} to our scenario). Moreover, we would also like to explore non-greedy clustering techniques such as Affinity Propagation.

\paragraph{Implementation Issues} Our current proposal does not scale well with the original (graph) signal dimension, as it mainly relies on an iterative pair-wise similarity computation between all (graph) elements. More research about how to relax this aspect is required in order to make its deployment feasible, and in this regard some ideas we want to explore are:
\begin{itemize}
    \item To test whether static higher-order structures generated from training samples can perform well in testing time; not only this would reduce the execution time, but also the necessity of storing the cluster sequence at each iteration.

    \item To check the feasibility of storing simultaneously sparse matrices indicating when the compression loss exceeds a certain threshold; apart from becoming a potential indicator of anomalies in the signal, if the distribution of big reconstruction errors is really sparse, combining them with the compressed representations will provide with loss tolerance guarantees, and could be key for matching \textit{zfp} performance.
    
    \item For large graphs, to divide the original graph into independent subgraphs, to which then our method is applied.
\end{itemize}

\paragraph{Topological MP} Another goal is to shed more light on the performance difference observed between our two architectures, SetMP and CombMP. Owing to current results, it seems that the lower complexity of SetMP is a clear advantage, and suggests that intermediate edge communications noisily interfere in the compression process. This should be further validated with other datasets, and possibly by testing some variations of the edge-based communication pipeline in the general CombMP architecture.

\end{document}